%% file: root.tex
\documentclass[]{l4dc2020} 

\title[Structured Variational Inference in Partially Observable Unstable GP-SSMs]{Structured Variational Inference in Partially Observable Unstable Gaussian Process State Space Models}
\usepackage{times}
\usepackage{mathtools}
\usepackage{booktabs}
\usepackage{tikz}
\usepackage{wrapfig}
\usepackage{cleveref}
\crefname{equation}{eq.}{equations}
\Crefname{equation}{Eq.}{Equations}
\crefname{table}{table}{tables}
\Crefname{table}{Table}{Tables}

\crefname{figure}{fig.}{figures}
\Crefname{figure}{Fig.}{Figures}

\DeclarePairedDelimiterX{\infdivx}[2]{(}{)}{%
  #1\;\delimsize|\delimsize|\;#2%
}
\newcommand{\cbfssm}{\textsc{CBF-SSM}\xspace}
\newcommand{\prssm}{\textsc{PR-SSM}\xspace}
\newcommand{\vcdt}{\textsc{VCDT}\xspace}
\newcommand{\KL}[2]{\text{KL}\infdivx{#1}{#2}}

\newcommand*\samethanks[1][\value{footnote}]{\footnotemark[#1]}
\hypersetup{
    colorlinks = false,
}% Use \Name{Author Name} to specify the name.
\author{%
 \Name{Silvan Melchior}\thanks{equal contribution} \Email{silvan.melchior@alumni.ethz.ch}\\
 \Name{Sebastian Curi}\samethanks \Email{sebastian.curi@inf.ethz.ch} \\
 \Name{Felix Berkenkamp} \Email{befelix@inf.ethz.ch}\\
 \Name{Andreas Krause} \Email{krausea@inf.ethz.ch}\\
 \addr Department of Computer Science, ETH Zurich%
}

\begin{document}

\maketitle

\input{sections/abstract}


\input{sections/introduction}
\input{sections/related_work}
\input{sections/problem_statement}
\input{sections/cbf_ssm}
\input{sections/experiments}
\input{sections/conclusions}

\acks{
This project has received funding from the European Research Council (ERC) under the European Union’s Horizon 2020 research and innovation programme grant agreement No 815943.
It was also supported by a fellowship from the Open Philanthropy Project.
We would like to thank Karen Bodie and Maximilian Brunner for the Voliro robot data and valuable discussions.
}

\bibliography{references}

\end{document}

%% file: sections/abstract.tex
%!TEX root = ../root.tex

\begin{abstract}
We propose a new variational inference algorithm for learning in Gaussian Process State-Space Models (GPSSMs).
Our algorithm enables learning of unstable and partially observable systems, where previous algorithms fail.
Our main algorithmic contribution is a novel approximate posterior that can be calculated efficiently using a single forward and backward pass along the training trajectories.
The forward-backward pass is inspired on Kalman smoothing for linear dynamical systems but generalizes to GPSSMs.
Our second contribution is a modification of the conditioning step that effectively lowers the Kalman gain.
This modification is crucial to attaining good \emph{test} performance where no measurements are available.
Finally, we show experimentally that our learning algorithm performs well in stable and unstable \emph{real} systems with hidden states.
\end{abstract}

%% file: sections/introduction.tex
%!TEX root = ../root.tex

\section{Introduction}
% Motivations for the research problem
We consider the problem of learning a probabilistic model of a non-linear dynamical system from data as a first-step of model-based reinforcement learning \citep{berkenkamp2019safe,kamthe2017}.
High-stake control applications require the model to have great predictive performance \emph{in expectation} as well as a correct \textit{uncertainty quantification} over all the prediction sequence.
Although parametric models such as deep neural networks successfully achieve the former \citep{chua2018deep,archer2015black}, they do not provide correct probability estimates \citep{guo2017calibration,malik2019calibrated}.
Instead, we consider Gaussian Processes-State Space Models (GP-SSMs), which were introduced by \citet{wang2006gaussian}.
These models meet both requirements at the cost of computationally costlier predictions and involved inference methods \citep[Section 3.4]{ialongo2019overcoming}.
%(require $O(T^3)$ computation to predict a trajectory of length $T$) .

% What is the problem with the state of the art that we're solving.
State-of-the-Art inference methods on GP-SSMs models use doubly stochastic variational inference \citep{salimbeni2017} on proposed approximate posteriors that are \emph{easy} to sample.
The \prssm algorithm, by \citet{doerr2018}, uses an approximate posterior that preserves the predictive temporal correlations of the prior distribution.
\prssm has great test performance in some tasks but in others it fails to learn the system.
\citet{ialongo2019overcoming} address \prssm limitations and propose an approximate posterior that conditions on measurements using Kalman Filtering \citep{kalman1960}, leading to the \vcdt algorithm.
Although \vcdt gives accurate predictions in cases where \prssm fails, it has worse performance in tasks where \prssm successfully learns the system.
Furthermore, there are tasks in which both algorithms fail to learn dynamical systems.

\begin{figure}
    \centering
    \begin{subfigure}[\prssm]{
        \includegraphics[width=0.21\textwidth]{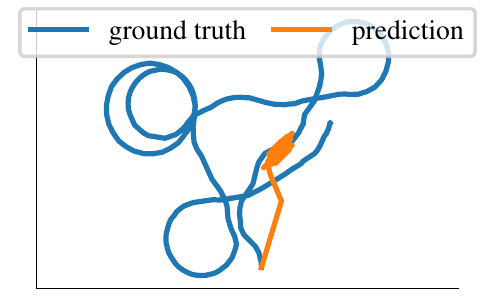}
        \label{sfig:PRSSM_robomove_simple}
    }
    \end{subfigure}
    \begin{subfigure}[\vcdt-Full]{
        \includegraphics[width=0.21\textwidth]{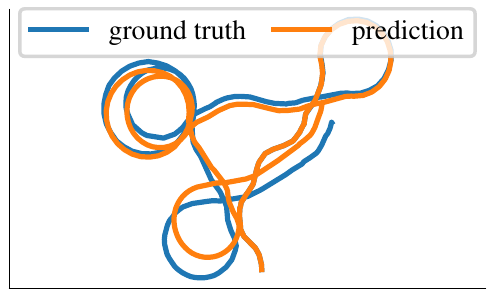}
        \label{sfig:VCDT_robomove_simple}
    }
    \end{subfigure} 
    \begin{subfigure}[\vcdt-Partial]{
        \includegraphics[width=0.21\textwidth]{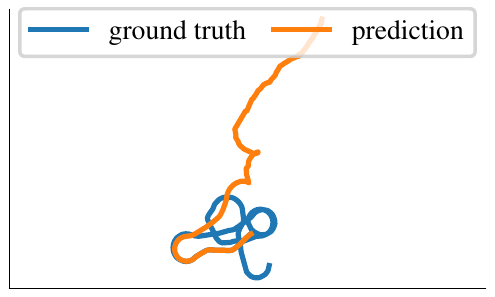}
        \label{sfig:VCDT_robomove}
    }
    \end{subfigure}
    \begin{subfigure}[\cbfssm]{
        \includegraphics[width=0.21\textwidth]{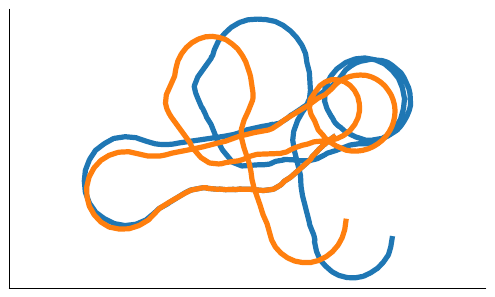}
        \label{sfig:CBFSSM_robomove}
    }
    \end{subfigure}
    
    \caption{Open-loop predictions on test set for a noisy Dubin's car model.
    In \Cref{sfig:PRSSM_robomove_simple,sfig:VCDT_robomove_simple} the full state is observed.
    \vcdt learns to predict correctly whereas \prssm explains observations with zero mean and high measurement noise.
    In \Cref{sfig:VCDT_robomove_simple,sfig:CBFSSM_robomove} only partial state information is available.
    \vcdt fails to account for partial observability, it overfits to the training set and the test-predictions diverge.
    \cbfssm instead use the smoother pass to infer the hidden states and it has good performance on the training set.} \vspace{-2em}
    \label{fig:robomove_mean_prediction}
\end{figure}

% Contributions / our solution
% Our first objective is to identify in which cases do \prssm or \vcdt algorithm fail and develop an algorithm that addresses these cases.
This paper builds on the observation that \prssm cannot learn systems that are not mean square stable (MSS) as the mismatch between the true and the approximate posterior can be arbitrarily large (\Cref{sfig:PRSSM_robomove_simple}).
Informally, a system is not MSS when the state uncertainty increases with time.
If the state is fully observed, \vcdt learns (\Cref{sfig:VCDT_robomove_simple}) as the conditioning step controls the uncertainty in the posterior.
However, when there are hidden states, \vcdt also fails (\Cref{sfig:VCDT_robomove}).
To address this issue, we introduce a backward smoother that is similar in spirit to the Kalman smoother.
We then condition using the smoothed estimates, instead of conditioning on the raw observations.
Our algorithm, Conditional Backward-Forward State Space Model (\cbfssm), succeeds in these tasks (\Cref{sfig:CBFSSM_robomove}) and reduces to \vcdt when full state information is available.
%A drawback of \vcdt is that it does not achieve the same performance as \prssm in MSS tasks.
The second improvement of our algorithm is that we reduce the Kalman gain in the conditioning step.
This is crucial to achieve good \emph{test} predictive performance, where no measurements are available.
We parametrize the conditioning level with a single parameter $k$ that explicitly interpolates between the full conditioning (as in \vcdt) and no conditioning (as in \prssm) to achieve good performance in both MSS and not MSS tasks.

% Finally, we demonstrate with extensive experiments the success of \cbfssm.

%% file: sections/related_work.tex
%!TEX root = ../root.tex
\subsection{Related Work}
\paragraph{Variational Inference in GP-SSMs}
\citet{frigola2014} introduce variational inference in GP-SSMs using a mean-field approximation over the sequence of states.
To incorporate input-output measurements, 
\citet{mattos2016} introduce a recognition module that learns the initial state distribution.
\citet{eleftheriadis2017} overcome the mean-field approximation and propose a posterior that preserves the prior temporal correlations for \emph{linear} systems, while \citet{doerr2018} present a posterior that preserves the prior temporal correlations for \emph{non-linear} systems.
Finally, \citet{ialongo2019overcoming} approximate the posterior temporal correlation by conditioning the prior on a single observation (i.e.,~filtering).
We build upon these works and introduce a backward smoother used for conditioning that approximates the true posterior temporal correlations better than previous work.

\paragraph{Variational Inference on Parametric State Space Models}
\citet{archer2015black} introduce stochastic variational inference on parametric state-space models using a Gaussian distribution with a structured covariance matrix to obtain a tractable posterior.
\citet{krishnan2017structured} build on this work relaxing the structure of the covariance matrix and introducing a deterministic smoothing pass.
Our backward pass is similar in spirit, but we consider probabilistic smoothed observations instead of deterministic ones to account for uncertainty in the backward pass explicitly.

%% file: sections/problem_statement.tex
%!TEX root = ../root.tex
\section{Problem Statement and Background}
We consider the problem of model-learning: At test time, we are given a sequence of control actions $u_{1:T}$ together with initial observations $y_{1:t'}$ and we must predict future observations $y_{t':T}$.
We need an initial sequence $t'$ of observations as the initial state is hidden, i.e., $t'$ is the system lag \citep{markovsky2008data}.
During training, we have access to a training data set that consists of sequences of actions and corresponding observations.
We evaluate the quality of our model by evaluating the log-likelihood of the true observations and the RMSE of the mean predictions.
%In this section, we introduce GP-SSMs and review the variational inference methods by~\citet{doerr2018} and \citet{ialongo2019overcoming} that our paper builds on.

\paragraph{Gaussian Process}
A Gaussian Process (GP) is a distribution over functions $f \colon \mathbb{R}^{d_x} \to \mathbb{R}$ that is parametrized by a mean function $m(\cdot)$ and covariance function $k(\cdot,\cdot)$, which respectively encode the expected value and similarities in the input space.
%Any finite collection of function values $\mathbf{f} = \{f(x_1), \dots, f(x_N) \}$ at arbitrary inputs $\mathbf{x} = \{x_1, \dots, x_N\}$ will have a joint gaussian distribution, with mean vector with $i$-th entry $\mu_{i} = m(x_i)$ and covariance matrix with $i,j$-th entry $\Sigma_{i, j} = k(x_i, x_j)$.
Given a prior $f \sim \mathcal{GP}(m(\cdot), k(\cdot, \cdot))$ and observations $(x , f_x )$, the posterior distribution of $f$ is also a GP with mean and covariance at $x'$
\begin{equation}
  \mu(x') = m(x') + k_{x'} K^{-1}_{x, x} (f_x - m_x),
  \hspace{1em}
  \Sigma(x', x') = k(x', x) - k_{x'}^\top K^{-1}_{x, x} k_{x'},
  \label{eq:gp_predictive}
\end{equation}
where $m_x = \{ m(x_1), \dots, m(x_n) \}$, $f_x = f(x)$, $[k_{x'}]_{i, j} = k(x'_i, x_j)$ and $[K_{x, x}]_{i,j} = k(x_i, x_j)$.

%For multi-dimensional transition functions $f$ with $d_x>1$ we use independent GPs for each dimension to reduce computational complexity, although our method is not limited to this choice.

\paragraph{Gaussian Process State-Space Model}

We model the process that generates observations with a SSM.
The Markovian latent state $x \in \mathbb{R}^{d_x}$ evolves over time based on a transition function $f$.
The key aspect of these models is that we place a GP prior on these functions.
At every time step $t$, we obtain measurements $y_t \in \mathbb{R}^{d_y}$ of the state~$x_t$.
The state transitions and observations are corrupted by zero-mean Gaussian noise with covariance matrices $\Sigma_x$ and $\Sigma_y$, respectively.
The GP-SSM is
\begin{equation}
  f \sim \mathcal{GP}(m(\cdot), k(\cdot, \cdot)), \hspace{.5em} 
  x_1 \sim \mathcal{N}(\mu_1, \Sigma_1), \hspace{.5em}
  x_{t+1} | f_t, x_t \sim \mathcal{N}(f(x_t), \Sigma_x), \hspace{.5em}
  y_t \sim \mathcal{N} (C x_t, \Sigma_y).
  \label{eq:state_space_model}
\end{equation}

For multi-dimensional transition functions $f$ with $d_x>1$, we use independent GPs for each dimension to reduce computational complexity, although our method is not limited to this choice.
Furthermore, we restrict $C=\begin{bmatrix}I & 0 \end{bmatrix}$, and $\Sigma_x$ and $\Sigma_y$ to be diagonal to capture the correlations between the states components only through $f$.
For brevity, we omit control inputs.
However, all derivations extend to controlled systems and the systems in the experiments have controls.

\paragraph{Sparse GP Approximation}
The memory needed to compute the approximate posterior of a GP for $N$ observations scales as $\mathcal{O}(N^2)$ and the computational complexity as $\mathcal{O}(N^3)$.
These requirements make GPs intractable for large-scale problems.
Furthermore, the GP model \eqref{eq:gp_predictive} assumes that the inputs are deterministic, whereas the inputs to the GP in model \eqref{eq:state_space_model} are probabilistic.
To address both issues we use sparse GPs \citep{titsias2009variational,hensman2013gaussian}.
In such models, the GP specifies function values $u_f$ at $M$ input locations $z_f$ such that $p(u_f) = \mathcal{N} (\mu^{u_f}, \Sigma^{u_f})$.
The function value at a location $x'$ different to $z_f$ follows a distribution given by $f(x') \sim \int p(f(x') | u_f) p(u_f) \mathrm{d}u_f$, where $ p(f(x') | u_f)$ is the posterior of $f$ at location $x'$ given pseudo-observations $(z_f, u_f)$ (see \Cref{eq:gp_predictive}).
Hence, $f(x')$ is Gaussian and can be computed in closed form.
When $M \ll N$, this brings a large computational advantage and does not require the true inputs $x$ to be deterministic.
The sparse GP-SSM prior and posterior distribution are
\begin{subequations}
\begin{align}
    p(u_f, x_{1:T}, y_{1:T}) &= p(x_1) p(u_f) \prod\nolimits_{t=1}^{T-1} p(x_{t+1} \mid f_{t}, x_{t}) p(f_{t} \mid u_f) \prod\nolimits_{t=1}^T p(y_t \mid x_t), \label{eq:prior} \\
    p(u_f, x_{1:T} \mid y_{1:T}) &= p(x_1 \mid y_{1:T}) p(u_f \mid y_{1:T}) \prod_{t=1}^{T-1} p(x_{t+1} \mid x_{t}, f_{t}, y_{t+1:T}) p(f_{t} \mid u_f, y_{1:T}). \label{eq:posterior}
\end{align}
\end{subequations}
\paragraph{Prediction with GPSSMs}
The model \eqref{eq:state_space_model} specifies a mechanism to generate samples from the GPSSM.
For the trajectory to be consistent, the function sampled along the trajectory has to be unique.
To ensure this for a trajectory of length $T$, we need to condition on all the previous observations yielding a computational complexity of $O(T^3)$.
\citet{doerr2018} omit the consistency requirement and uses independent samples of $f$ for each time-step prediction by assuming that $\int p(u_f) \prod_{t=2}^T p(f_{t-1} \mid u_f) \mathrm{d}u_f = \prod_{t=2}^T \int p(u_f) p(f_{t-1} \mid u_f)\mathrm{d}u_f$, i.e., each transition is independent of each other.
\citet{ialongo2019overcoming} criticizes this assumption and instead proposes to sample $u_f \sim p(u_f)$ at the beginning of each trajectory and approximate the integral by using a Monte Carlo approximation.
\citet{mchutchon2015nonlinear} also addresses the cubic sampling by using just the mean of $p(u_f)$ in each trajectory.
Another possibility is to degenerate $p(u_f)$ to a delta distribution in which all methods coincide but essentially reduces the model to a parametric one.

\paragraph{Learning in GPSSMs}
The posterior distribution \eqref{eq:posterior} is intractable to compute when the transitions are non-linear.
Traditional methods such as MCMC \citep{frigola2013} do not scale to large datasets.
Variational inference methods \citep{blei2017} propose an approximate posterior $q(u_f, x_{1:T}, y_{1:T})$ that is easy to sample and minimize the KL divergence between the approximate and the true posterior.
This procedure turns out to be equivalent to maximizing the evidence lower bound (ELBO).
The approximate posterior of \prssm and \vcdt are
\begin{subequations}
\begin{align}
    q_{\prssm}(u_f, x_{1:T}, y_{1:T}) &= q(x_1|y_{1:t'}) q(u_f) \prod\nolimits_{t=1}^{T-1} p(x_{t+1} | x_{t}, f_{t}) p(f_t|u_f), \label{eq:PRSSM-posterior} \\
    q_{\vcdt}(u_f, x_{1:T}, y_{1:T}) &= q(x_1|y_{1:t'}) q(u_f) \prod\nolimits_{t=1}^{T-1} q(x_{t+1} | x_{t}, f_{t}, y_{t+1}) p(f_t|u_f), \label{eq:VCDT-posterior}
\end{align} \label{eq:approx-posterior}
\end{subequations}
where $q(x_1 \mid y_{1:t'}) = \mathcal{N}(\mu^{qx_1}, \Sigma^{qx_1})$ is called the recognition module and $q(u_f)= \mathcal{N}(\mu^{qu_{f}}, \Sigma^{qu_f})$ is the sparse GP posterior.
Both algorithms use the prior $p(f_t|u_f)$ to generate the function samples which simplifies the KL divergence between the function prior and posterior to the KL divergence between $q(u_f)$ and $p(u_f)$ only \citep{matthews2017scalable}.
The crucial difference between both algorithms is on how they compute the next-state approximate posterior.
Whereas \prssm uses the prior, \vcdt uses a 1-step approximation to the posterior (c.f.~ \Cref{eq:prior,eq:posterior}).
The 1-step \vcdt posterior approximation is also a Gaussian that can be efficiently computed using a Kalman-filtering conditioning rule.
The ELBO of \prssm and \vcdt are
\begin{subequations}
\begin{align}
    \mathcal{L}_\prssm &= \sum\nolimits_{t=1}^T \mathbb{E}_q \left[\log p(y_t|x_t)\right] - \KL{q(u_f)}{p(u_f)} - \KL{q(x_1|y_{1:t'})}{p(x_1)}, \label{eq:PRSSM-elbo} \\
    \mathcal{L}_\vcdt &= \mathcal{L}_\prssm - \sum\nolimits_{t=1}^{T-1} \KL{q(x_{t+1}|x_{t}, f_{t}, y_{t+1})}{p(x_{t+1} \mid f_{t}, x_{t}) }. \label{eq:VCDT-elbo} 
\end{align} 
\label{eq:ELBO}
\end{subequations}
The first term of the ELBO \eqref{eq:PRSSM-elbo} maximizes the observations conditional likelihood, whereas the first KL divergence term regularizes the inducing points of the GPs and the recognition module. It is common to select $p(x_1)$ as an uninformative prior, so this KL divergence vanishes. The ELBO of \vcdt \eqref{eq:VCDT-elbo} also regularizes the conditioning step through the KL divergence. 

%% file: sections/cbf_ssm.tex
%!TEX root = ../root.tex

\section{Variational Inference in Unstable GP-SSMs} \label{sec:CBFSSM}

\paragraph{Mean-Square Unstable Systems} A system that is mean-square stable (MSS) has a bounded predictive state covariance matrix $\lim_{t \to \infty} \mathbb{E}\left[x_t x_t^\top \mid x_1 \right]$ \citep{soong1973,Khasminskii2012Stochastic-1}.
Conversely, systems that are not MSS have an unbounded predictive state covariance matrix.
A linear system with a spectral radius larger or equal to one, combined with non-zero additive noise, is not MSS.
As an illustrative example, we use Dubin's car model as a not MSS system, where the state is the $(x, y)$ position and the orientation, and the controls are the speed and curvature commands.

Learning with \prssm on not MSS systems over long-time horizons is challenging because the state-transition term in the approximate posterior \eqref{eq:approx-posterior} does not condition on the observations as the true posterior \eqref{eq:posterior} does.
In such models, the approximate posterior variance increases along the trajectory, whereas the true posterior variance is constant.
When optimizing the ELBO \eqref{eq:PRSSM-elbo}, the model assigns high observation noise $\Sigma_y$ to explain the measurements instead of learning $f$.

When the sequence is \emph{short}-enough, \prssm does not suffer this shortcoming during training, but the test performance on long sequences is poor.
\vcdt addresses this by using an approximate posterior that conditions on the measurements.
Nevertheless, it learns to condition \emph{too much} on the observations, which are not present during testing leading to poor performance.
Furthermore, when the system has \emph{unobserved states}, the conditioning step only corrects the measured components of the state.
In contrast, the unmeasured ones are given by the prior distribution as in \prssm.
The Conditional Backward-Forward State-Space Model (\cbfssm) algorithm explicitly estimates the hidden states and learns even with partial state observation and in unstable systems.

\subsection{Conditional Backward-Forward State-Space Model}
Ideally, we would like to propose an approximate posterior that uses the full $y_{t:T}$ in the conditional state transition term, yet it is tractable to compute.
We propose a backward pass to smooth the measurements $y_{t:T}$ into a distribution over a single pseudo-state $\tilde{x}_t \in \mathbb{R}^{d_x}$ that approximates $p(\tilde{x}_t | y_{t:T})$.

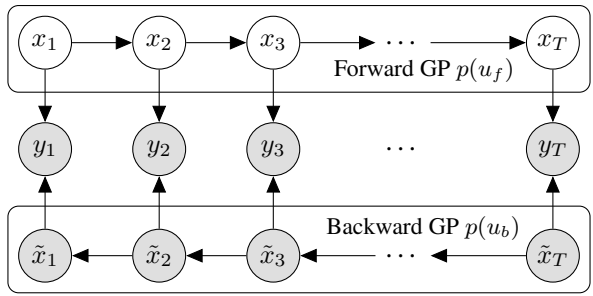
\begin{wrapfigure}{r}{8.0cm} \vspace{-1em}
 \input{figures/graphical_model}
\vspace{-0.5em}
 \caption{Backward-Forward GP-SSM Model}
 \label{fig:CBFSSM}
\end{wrapfigure}
\paragraph{Backward Pass} Traditional smoothing algorithms compute the posterior by propagating $p(x_t \mid x_{t+1}, y_t)$ in a backward pass.
However, when the forward model has a Gaussian Process prior, the backward probabilities are intractable.
Instead, we propose an auxiliary noiseless model that runs from $t=T$ to $t=1$ that produces the same observations $y_t$, as shown in \Cref{fig:CBFSSM}.
This model has states $\tilde{x}_t \in \mathbb{R}^{d_x}$ and is generated as
\begin{equation}
  f_b \sim \mathcal{GP}(m(\cdot), k(\cdot, \cdot)), \hspace{.5em} 
  \tilde{x}_T \sim \mathcal{N}(\mu_1, \Sigma_1), \hspace{.5em}
  \tilde{x}_{t} = f_b(\tilde{x}_{t+1}), \hspace{.5em}
  y_t = C \tilde{x}_t.
  \label{eq:backward_state_space_model}
\end{equation}
Using a sparse GP approximation for the backward pass, the \cbfssm approximate posterior is:
\begin{align}
    q_{\cbfssm}(u_f, x_{1:T}, y_{1:T}) &= q(x_1\mid y_{1:t'}) q(u_f) \prod\nolimits_{t=1}^{T-1} q(x_{t+1} \mid x_{t}, f_{t}, \tilde{x}_{t+1}) p(f_t\mid u_f) \nonumber \\
    & \quad \cdot q(\tilde{x}_T \mid y_T) q(u_b) \prod\nolimits_{t=1}^{T-1} p(\tilde{x}_{t} \mid \tilde{x}_{t+1}, f_{t+1}, y_t) p(f_t\mid u_b).
    \label{eq:CBFSSM-posterior}
\end{align}
The second line of \Cref{eq:CBFSSM-posterior} is computed with a single backward pass and the first line with a single forward pass, conditioning on $\tilde{x}_t$ at every time step.
The first $d_y$ components of $\tilde{x}_t$ are $y_t$ and the rest are predicted with the backward GP.
When the state is fully observed, the second line of \Cref{eq:CBFSSM-posterior} reduces to a dirac distribution at $\tilde{x}_t = y_t$ and \cbfssm and \vcdt algorithms coincide.
This forward-backward algorithm is similar in spirit to the smoother from \citet{krishnan2017structured}, but our models are probabilistic to approximate the true posterior.
The ELBO of \cbfssm is
\begin{equation*}
   \mathcal{L}_\cbfssm = \mathcal{L}_\prssm - \hspace{-0.1em} \sum\nolimits_{t=1}^{T-1} \KL{q(x_{t+1}|x_{t}, f_{t}, \tilde{x}_{t+1})}{p(x_{t+1} \mid f_{t}, x_{t}) } - \KL{q(u_b)}{p(u_b)}. %\label{eq:CBFSSM-elbo}
\end{equation*}

\paragraph{Soft Conditioning Step} The conditioning step of \vcdt for full state observations can be summarized as follows.
As both $q(x_t \mid f_{t-1}, x_{t-1}) \equiv \mathcal{N}(\mu^{-}_t, \Sigma^{-}_t)$ and $p(\tilde{x}_t \mid x_t) \equiv \mathcal{N}(x_t, \tilde{\Sigma}_x)$ are Gaussian distributions, the approximate posterior $q(x_{t} \mid f_{t-1}, x_{t-1}, \tilde{x}_{t}) = \mathcal{N}(\mu_t, \Sigma_t)$ with
\begin{equation}
 \mu_t = \mu^{-}_t + K(\tilde{y}_t - \mu_t^{-1}), \hspace{1em} \Sigma_t = (I - K) \Sigma^{-}_t (I-K)^\top + K \tilde{\Sigma}_x K^\top, \hspace{1em} \label{eq:conditioning}
\end{equation}
where $K$ is the Kalman gain $K = \Sigma^{-}_t (\tilde{\Sigma}_x + \Sigma^{-}_t)^{-1}$.
Our second contribution is a soft conditioning step.
We propose to use a free factor $k \geq 1$ such that the Kalman gain is $K_{\text{soft}} = \Sigma^{-}_t ( \tilde{\Sigma}_x + k \Sigma^{-}_t)^{-1}$ and the conditioning step is still given by \Cref{eq:conditioning}.
When $k=1$, this reduces to the \vcdt conditioning step and, when $k \to \infty$ then $K_{\text{soft}} \to 0$, and \cbfssm does not condition, as in \prssm.
The soft-conditioning parameter $k$ trades off one-step and long-term accuracy.
This soft-conditioning step is a particular case of the most general posterior proposed by \citet{ialongo2019overcoming}.
However, their function class is time-varying and much larger than our restricted soft-conditioning step.
Hence, \vcdt tends to overfit and produce poor test results, as we found in experiments.

\paragraph{Tuning Hyper-Parameters} In standard stochastic variational inference \citep{hoffman2013}, the KL-divergence terms are re-weighted by factors to account for batch-size relative to the full dataset.
In our setting, the i.i.d.~assumption of the dataset is violated, and this leads to sub-optimal results in all three algorithms.
We introduce a scaling parameter $\beta$ to reweigh the KL-divergence terms in the ELBO.
This re-weighting scheme is based on the $\beta$-VAE algorithm by \citet{higgins2017beta}.
Furthermore, we notice that when sampling independent functions along a trajectory as in \prssm, the KL divergence of the inducing points has to be scaled by the trajectory length.

%% file: figures/graphical_model.tex
\usetikzlibrary{bayesnet}
\begin{tikzpicture}
  \node[latent]                               (x1) {$x_1$};
  \node[latent, right=0.8 cm of x1]           (x2) {$x_2$};
  \node[latent, right=0.8 cm of x2]           (x3) {$x_3$};
  \node[right=1. cm of x3]                    (xn) {\ldots};
  \node[latent, right=3.0 cm of x3]           (xt) {$x_T$};
  {
  \tikzset{plate caption/.append style={right=4.2cm of #1.south west}}
  \plate{Forward} {(x1) (x2) (x3) (xn) (xt)} {Forward GP $p(u_f)$};
  }

  \node[obs, below=.7 cm of x1]                (y1) {$y_1$};
  \node[obs, below=.7 cm of x2]                (y2) {$y_2$};
  \node[obs, below=.7 cm of x3]                (y3) {$y_3$};
  \node[right=1. cm of y3]                    (yn) {\ldots};
  \node[obs, below=.7 cm of xt]                (yt) {$y_T$};

  \node[obs, below=.7 cm of y1]                (x1_) {$\tilde{x}_1$};
  \node[obs, below=.7 cm of y2]                (x2_) {$\tilde{x}_2$};
  \node[obs, below=.7 cm of y3]                (x3_) {$\tilde{x}_3$};
  \node[right=1. cm of x3_]                   (xn_) {\ldots};
  \node[obs, below=.7 cm of yt]                (xt_) {$\tilde{x}_T$};
  
  {
  \tikzset{plate caption/.append style={right=4.1cm of #1.north west}}
  \plate{Backward} {(x1_) (x2_) (x3_) (xn_) (xt_)} {Backward GP $p(u_b)$};
  }
  
    \edge{x1}{x2};
    \edge{x2}{x3};
    \edge{x3}{xn};
    \edge{xn}{xt};
    
    \edge{x1}{y1};
    \edge{x2}{y2};
    \edge{x3}{y3};
    \edge{xt}{yt};
    
    \edge{x2_}{x1_};
    \edge{x3_}{x2_};
    \edge{xn_}{x3_};
    \edge{xt_}{xn_};
    
    \edge{x1_}{y1};
    \edge{x2_}{y2};
    \edge{x3_}{y3};
    \edge{xt_}{yt};
\end{tikzpicture}

%% file: sections/experiments.tex
%!TEX root = ../root.tex

\section{Experiments}

\paragraph{System Identification Benchmarks}
We compare \cbfssm against \prssm and \vcdt on the datasets used by \citet{doerr2018}, where \prssm outperforms other methods.
\Cref{tab:rmse_smallscale} shows \cbfssm-1 without soft conditioning, \cbfssm-50 with a soft conditioning factor of $k=50$ and \cbfssm-1S without soft conditioning but with the function sampling method proposed by \citet{ialongo2019overcoming}.
We first remark that our implementation of \prssm has better performance than the original paper, and this is because we correctly compute the KL divergence between the inducing points when the functions are sampled independently along a trajectory.
The second observation is that \vcdt performs considerably worse than \prssm in these tasks.
If we compare \vcdt to \cbfssm-1 (both methods coincide except for the function sampling method and the backward pass), we see that \cbfssm-1 outperforms \vcdt.
If we compare \vcdt to \cbfssm-1S (both methods coincide except for the backward step), we see that the methods perform relatively similarly.
This suggests that the function sampling method proposed by \citet{ialongo2019overcoming} is too noisy to be useful for learning.
Finally, if we compare \cbfssm-1 to \cbfssm-50, we see that the performance is comparable, except for the large-scale Sarcos data set where soft conditioning is crucial to attaining good performance.
In summary, we see that \cbfssm-50 outperforms or is comparable to all other methods in all data sets.

\begin{table}[htpb]
\centering
\begin{tabular}{lccccc}
      \toprule
      Dataset           & \prssm & \vcdt & \cbfssm-1 & \cbfssm-50 & \cbfssm-1S \\
      \midrule
      \textsc{Actuator} & \textbf{0.446 (0.017)} & 1.060 (0.490) & \textbf{0.452 (0.014)} & \textbf{0.452 (0.013)} & 0.985 (0.360) \\
      \textsc{Ballbeam} & \textbf{0.045 (0.003)} & 0.080 (0.010) & \textbf{0.050 (0.004)} & \textbf{0.050 (0.004)} & 0.083 (0.007) \\
      \textsc{Drives}   & \textbf{0.332 (0.059)} & 0.757 (0.040) & \textbf{0.340 (0.073)} & \textbf{0.335 (0.068)} & 0.752 (0.019) \\
      \textsc{Dryer}    & \textbf{0.083 (0.005)} & 0.665 (0.230) & \textbf{0.080 (0.006)} & \textbf{0.082 (0.004)} & 0.650 (0.380) \\
      \textsc{Furnace}  & \textbf{1.260 (0.013)} & 3.020 (1.200) & \textbf{1.210 (0.018)} & \textbf{1.220 (0.010)} & 2.500 (0.640) \\
      \textsc{Flutter}  & \textbf{0.264 (0.012)} & 0.424 (0.060) & \textbf{0.273 (0.017)} & \textbf{0.269 (0.020)} & 0.714 (0.410) \\
      \textsc{Tank}     & 0.178 (0.007) & 0.846 (0.370) & \textbf{0.168 (0.013)} & \textbf{0.172 (0.014)} & 1.140 (0.340) \\
      \textsc{Sarcos}   & \textbf{0.046 (0.002)} & 0.170 (0.013) & 0.159 (0.012) & \textbf{0.049 (0.002)} & 0.265 (0.100) \\
      \bottomrule
    \end{tabular}
    \caption{Test RMSE [mean (std)] over five runs for different datasets.
    In bold typeface we indicate the best performing algorithms.
    \cbfssm-1 and \cbfssm-50 differ on the conditioning ($k=1$ or $k=50$).
    \cbfssm-1S ($k=1$) uses the \vcdt function sampling step.
    \cbfssm-50 achieves lowest error in all data sets. \vspace{-1em}}
    \label{tab:rmse_smallscale}
\end{table}

\begin{figure}[t]
  \centering
  \begin{subfigure}[Dubin's car model.]{
    \includegraphics[width=0.48\textwidth]{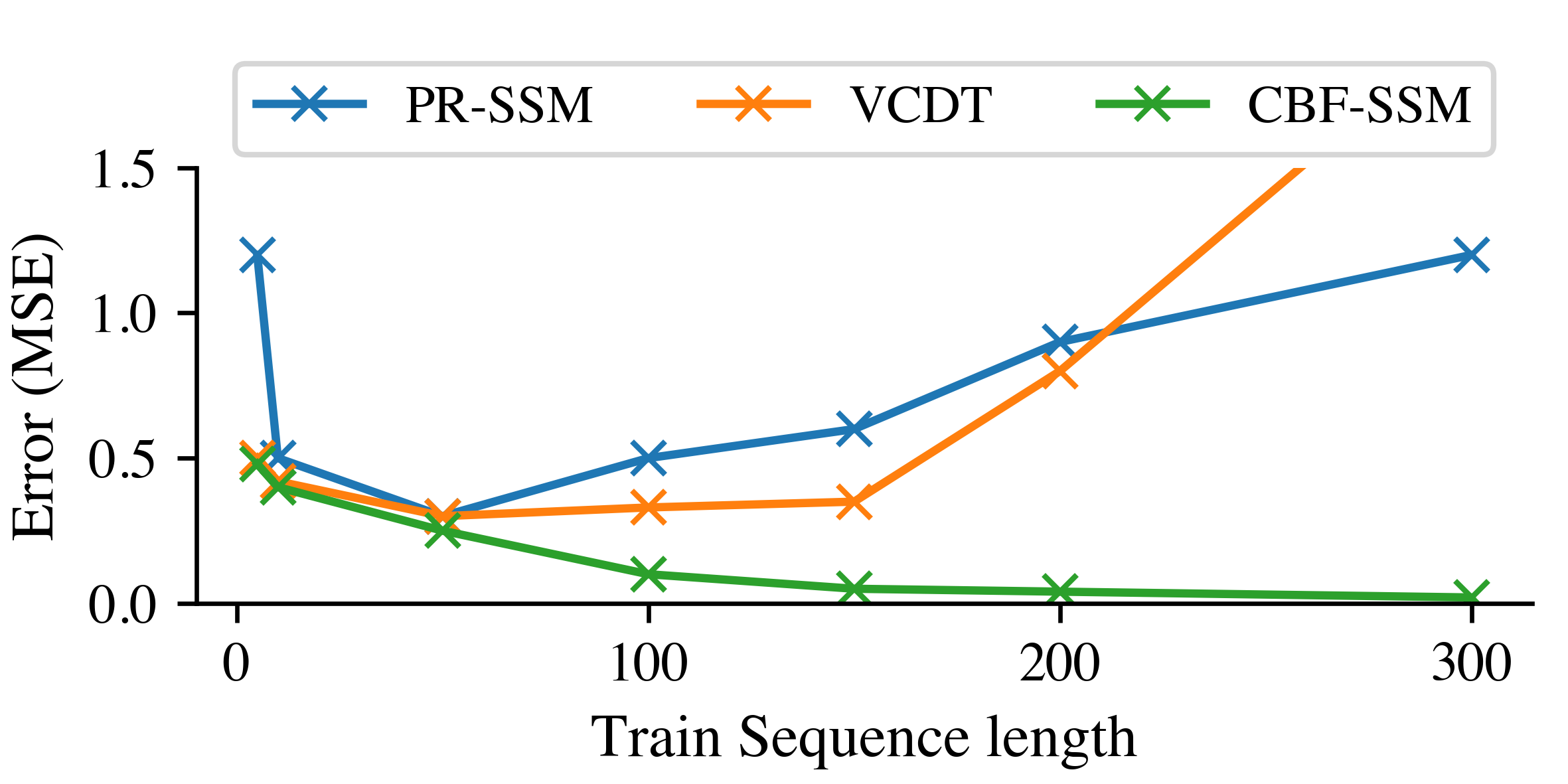}
    \label{fig:seqlen_graph}}
  \end{subfigure}%
  \begin{subfigure}[\textsc{VoliroX} drone (real-world).]{
    \includegraphics[width=0.48\textwidth,height=3.6cm]{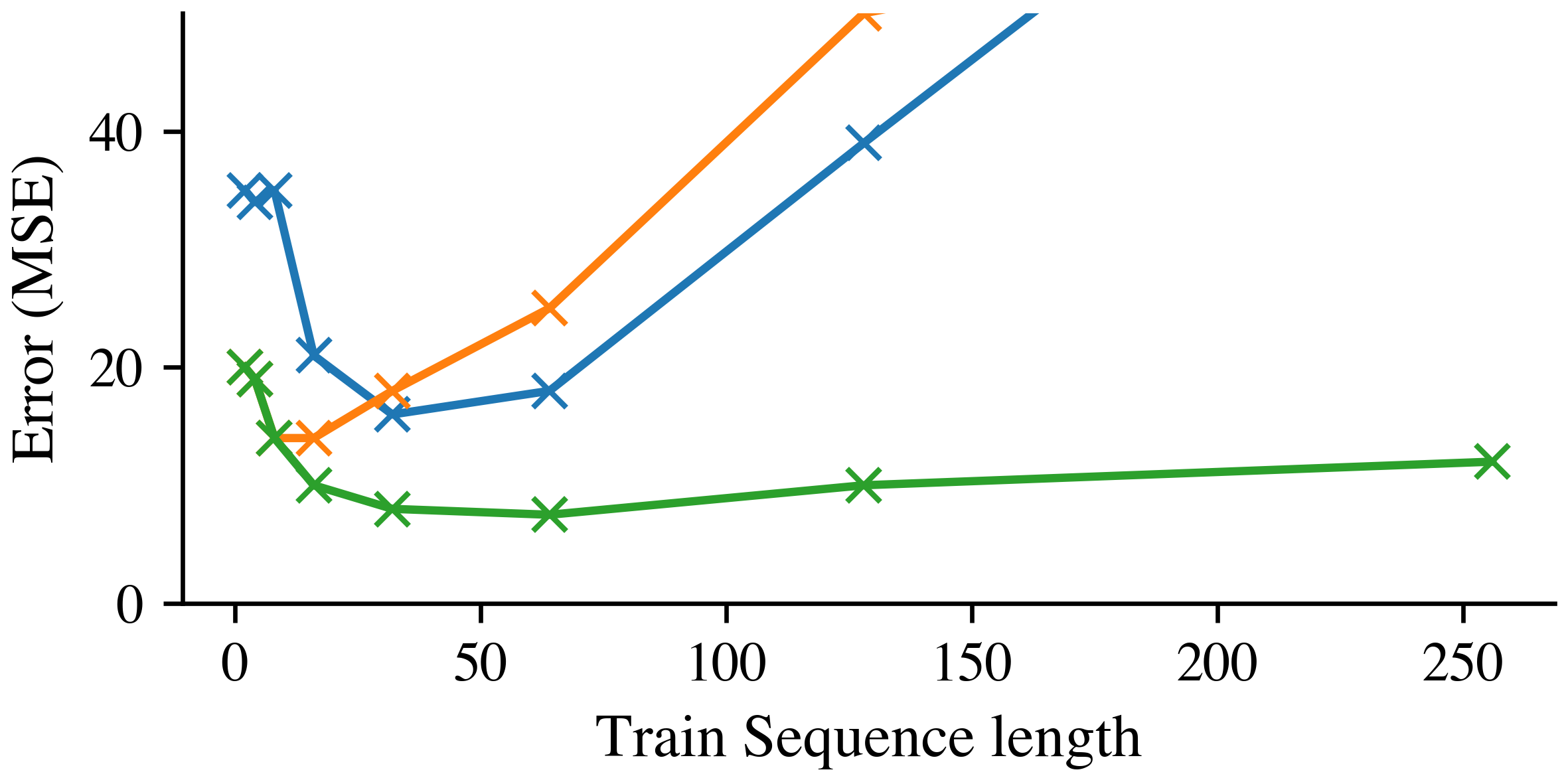}
    \label{fig:voliro_seqlen}}
  \end{subfigure}
  \caption{Effect of training sequence length on test error for MSU systems.
  \prssm can only train on short trajectories.
  \cbfssm achieves lower error by training on longer sequences. \vspace{-2em}}
\end{figure}

\paragraph{Simulated unstable system}
We evaluate the performance on the toy car dataset introduced in \Cref{sec:CBFSSM}.
\Cref{fig:robomove_mean_prediction} shows a qualitative comparison of the variational inference algorithms when trained on sequence lengths of 300 and the resulting test error for different sequence lengths is shown in \Cref{fig:seqlen_graph}.
\cbfssm achieves lower test error when training on longer sequences, while \prssm fails to learn the system accurately on long sequences.

\paragraph{VoliroX}
To demonstrate that \cbfssm can be applied to real-world, complex, and unstable systems, we use it to learn the dynamics of a flying robotic vehicle.
VoliroX \citep{bodie2018} is a drone consisting of twelve rotors mounted on six tiltable arms, which cause airflow interference that is difficult to model.
The dataset includes measured position and orientation $p \in \mathbb{R}^6$, while linear and angular velocities $v \in \mathbb{R}^6$ are unobserved.
Control inputs are the arm tilt angles $\alpha \in \mathbb{R}^6$ and motor signals $\eta \in \mathbb{R}^6$.
\citet{bodie2018} model the rigid body (RB) dynamics with an integrated ordinary differential equation (ODE), 
$(p_{t+1}, v_{t+1}) = f_{\text{RB-ODE}}(p_t, v_t, \xi_t, \tau_t)$, 
which depends on the forces $\xi_t$ and torques $\tau_t$ acting on the system.
While \citet{bodie2018} predict forces and torques with a physical model, $f_\mathrm{PM}$, we additionally learn a GP correction term to account for modeling errors, 
$(\xi_t, \tau_t) = f_{\mathrm{PM}}(\eta_t, \alpha_t) + f_{\mathrm{GP}}(\eta_t, \alpha_t)$.
We integrate the resulting ODE in a differentiable way using TensorFlow \citep{tensorflow2015-whitepaper} and estimate the velocities $\mathbf{v}$ with our backward model.
Although the system is high-dimensional, we use the GP only to model external forces and torques, $\mathbb{R}^{12} \to \mathbb{R}^{6}$.
Since we combine this prediction with the rigid body dynamics, we can effectively exploit prior physics knowledge and avoids learning about basic physics facts.

\begin{figure}[t]
    \begin{subfigure}[Physical Model only.]{
        \includegraphics[width=0.5\textwidth]{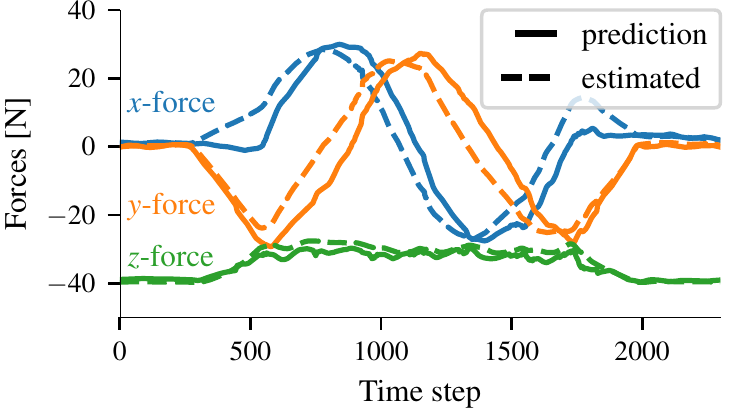}
        \label{sfig:voliro_physical}
        }
    \end{subfigure}
    \begin{subfigure}[Physical Model + \cbfssm.]{
        \includegraphics[width=0.44\textwidth]{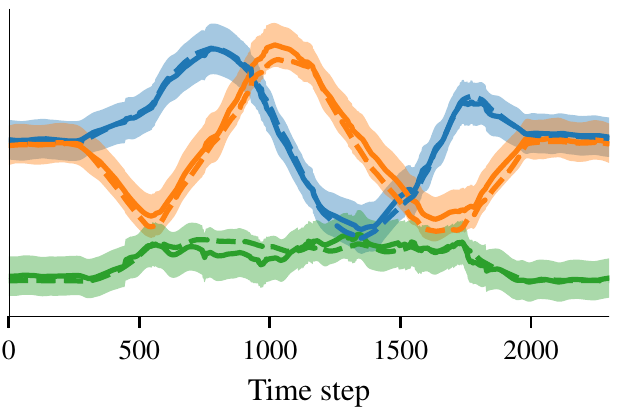}
        \label{sfig:voliro_cbfssm}
        }
    \end{subfigure}
    \caption{Test-set predictions on Voliro-X.
    In \Cref{sfig:voliro_physical} we show the forces predicted by the physical model and the forces estimated from data.
    In \Cref{sfig:voliro_cbfssm} we plot the predictions by \cbfssm.
    The shaded regions are $\pm 1.96$ the predicted std.~deviation. \vspace{-2em}}
\end{figure}

The physical model does not model airflow interference, which leads to significant prediction errors in \Cref{sfig:voliro_physical}.
In contrast, \cbfssm provides accurate predictions with reliable uncertainty information in \Cref{sfig:voliro_cbfssm}.
We compare these predictions to \prssm and \vcdt for different training sequence lengths in \Cref{fig:voliro_seqlen}.
Since the drone is unstable and has large process noise, \prssm and \vcdt can only train on short sequences.
In contrast, \cbfssm can reliably train on longer sequence lengths and hence achieve lower predictive errors without overfitting.
% Moreover, on the best sequence length for \prssm, \cbfssm achieves an MSE that is $47.2\%$ lower than that of \prssm.

\paragraph{Computational Performance} The prediction time of all algorithms is identical as all use the model \eqref{eq:state_space_model}. 
As a function of $T$, all algorithms take $\mathcal{O}(T)$ to compute the forward and the backward pass. 
However, the extra backward pass in our algorithm makes training $3.7\times$ slower. 

% \paragraph{Further Experiments} We provide extended experiments in \citet[Appendix C]{Melchior2019Structured}.

%% file: sections/conclusions.tex
%!TEX root = ../root.tex

\section{Conclusions}
We presented a new algorithm, \cbfssm, to learn on GPSSMs using Variational Inference.
Compared to previous work, our algorithm learns in both MSS and MSU systems with hidden states and achieves superior performance to all other algorithms.
We present two algorithmic innovations in \cbfssm: the backward pass that provides a better approximation to the true posterior and the soft conditioning that trades-off training and testing accuracy. Finally, we demonstrate the capabilities of \cbfssm in small and large-scale benchmarks and simulated and real robots.

%% file: root.bbl
\begin{thebibliography}{28}
\providecommand{\natexlab}[1]{#1}
\providecommand{\url}[1]{\texttt{#1}}
\expandafter\ifx\csname urlstyle\endcsname\relax
  \providecommand{\doi}[1]{doi: #1}\else
  \providecommand{\doi}{doi: \begingroup \urlstyle{rm}\Url}\fi

\bibitem[Abadi et~al.(2015)Abadi, Agarwal, Barham, Brevdo, Chen, Citro,
  Corrado, Davis, Dean, Devin, Ghemawat, Goodfellow, Harp, Irving, Isard, Jia,
  Jozefowicz, Kaiser, Kudlur, Levenberg, Man\'{e}, Monga, Moore, Murray, Olah,
  Schuster, Shlens, Steiner, Sutskever, Talwar, Tucker, Vanhoucke, Vasudevan,
  Vi\'{e}gas, Vinyals, Warden, Wattenberg, Wicke, Yu, and
  Zheng]{tensorflow2015-whitepaper}
Mart\'{\i}n Abadi, Ashish Agarwal, Paul Barham, Eugene Brevdo, Zhifeng Chen,
  Craig Citro, Greg~S. Corrado, Andy Davis, Jeffrey Dean, Matthieu Devin,
  Sanjay Ghemawat, Ian Goodfellow, Andrew Harp, Geoffrey Irving, Michael Isard,
  Yangqing Jia, Rafal Jozefowicz, Lukasz Kaiser, Manjunath Kudlur, Josh
  Levenberg, Dandelion Man\'{e}, Rajat Monga, Sherry Moore, Derek Murray, Chris
  Olah, Mike Schuster, Jonathon Shlens, Benoit Steiner, Ilya Sutskever, Kunal
  Talwar, Paul Tucker, Vincent Vanhoucke, Vijay Vasudevan, Fernanda Vi\'{e}gas,
  Oriol Vinyals, Pete Warden, Martin Wattenberg, Martin Wicke, Yuan Yu, and
  Xiaoqiang Zheng.
\newblock {TensorFlow}: Large-scale machine learning on heterogeneous systems,
  2015.
\newblock URL \url{https://www.tensorflow.org/}.
\newblock Software available from tensorflow.org.

\bibitem[Archer et~al.(2015)Archer, Park, Buesing, Cunningham, and
  Paninski]{archer2015black}
Evan Archer, Il~Memming Park, Lars Buesing, John Cunningham, and Liam Paninski.
\newblock Black box variational inference for state space models.
\newblock \emph{arXiv preprint arXiv:1511.07367}, 2015.

\bibitem[Berkenkamp(2019)]{berkenkamp2019safe}
Felix Berkenkamp.
\newblock \emph{Safe Exploration in Reinforcement Learning: Theory and
  Applications in Robotics}.
\newblock PhD thesis, ETH Zurich, 2019.

\bibitem[Blei et~al.(2017)Blei, Kucukelbir, and McAuliffe]{blei2017}
David~M Blei, Alp Kucukelbir, and Jon~D McAuliffe.
\newblock Variational inference: A review for statisticians.
\newblock \emph{Journal of the American Statistical Association}, 112\penalty0
  (518):\penalty0 859--877, 2017.

\bibitem[Bodie et~al.(2018)Bodie, Taylor, Kamel, and Siegwart]{bodie2018}
Karen Bodie, Zachary Taylor, Mina Kamel, and Roland Siegwart.
\newblock Towards efficient full pose omnidirectionality with overactuated
  mavs.
\newblock \emph{CoRR}, abs/1810.06258, 2018.

\bibitem[Chua et~al.(2018)Chua, Calandra, McAllister, and Levine]{chua2018deep}
Kurtland Chua, Roberto Calandra, Rowan McAllister, and Sergey Levine.
\newblock Deep reinforcement learning in a handful of trials using
  probabilistic dynamics models.
\newblock In \emph{Advances in Neural Information Processing Systems}, pages
  4754--4765, 2018.

\bibitem[Doerr et~al.(2018)Doerr, Daniel, Schiegg, Duy, Schaal, Toussaint, and
  Sebastian]{doerr2018}
Andreas Doerr, Christian Daniel, Martin Schiegg, Nguyen-Tuong Duy, Stefan
  Schaal, Marc Toussaint, and Trimpe Sebastian.
\newblock Probabilistic recurrent state-space models.
\newblock In \emph{Proceedings of the 35th International Conference on Machine
  Learning}, pages 1280--1289, 2018.

\bibitem[Eleftheriadis et~al.(2017)Eleftheriadis, Nicholson, Deisenroth, and
  Hensman]{eleftheriadis2017}
Stefanos Eleftheriadis, Tom Nicholson, Marc Deisenroth, and James Hensman.
\newblock Identification of gaussian process state space models.
\newblock In \emph{Advances in Neural Information Processing Systems}, pages
  5309--5319, 2017.

\bibitem[Frigola et~al.(2013)Frigola, Lindsten, Sch{\"o}n, and
  Rasmussen]{frigola2013}
Roger Frigola, Fredrik Lindsten, Thomas~B Sch{\"o}n, and Carl~Edward Rasmussen.
\newblock Bayesian inference and learning in gaussian process state-space
  models with particle mcmc.
\newblock In \emph{Advances in Neural Information Processing Systems}, pages
  3156--3164, 2013.

\bibitem[Frigola et~al.(2014)Frigola, Chen, and Rasmussen]{frigola2014}
Roger Frigola, Yutian Chen, and Carl~Edward Rasmussen.
\newblock Variational gaussian process state-space models.
\newblock In \emph{Advances in Neural Information Processing Systems}, pages
  3680--3688, 2014.

\bibitem[Guo et~al.(2017)Guo, Pleiss, Sun, and Weinberger]{guo2017calibration}
Chuan Guo, Geoff Pleiss, Yu~Sun, and Kilian~Q Weinberger.
\newblock On calibration of modern neural networks.
\newblock In \emph{Proceedings of the 34th International Conference on Machine
  Learning-Volume 70}, pages 1321--1330. JMLR. org, 2017.

\bibitem[Hensman et~al.(2013)Hensman, Fusi, and Lawrence]{hensman2013gaussian}
James Hensman, Nicolo Fusi, and Neil~D Lawrence.
\newblock Gaussian processes for big data.
\newblock In \emph{Uncertainty in Artificial Intelligence}, page 282. Citeseer,
  2013.

\bibitem[Higgins et~al.(2017)Higgins, Matthey, Pal, Burgess, Glorot, Botvinick,
  Mohamed, and Lerchner]{higgins2017beta}
Irina Higgins, Loic Matthey, Arka Pal, Christopher Burgess, Xavier Glorot,
  Matthew Botvinick, Shakir Mohamed, and Alexander Lerchner.
\newblock beta-vae: Learning basic visual concepts with a constrained
  variational framework.
\newblock \emph{ICLR}, 2\penalty0 (5):\penalty0 6, 2017.

\bibitem[Hoffman et~al.(2013)Hoffman, Blei, Wang, and Paisley]{hoffman2013}
Matthew~D Hoffman, David~M Blei, Chong Wang, and John Paisley.
\newblock Stochastic variational inference.
\newblock \emph{The Journal of Machine Learning Research}, 14\penalty0
  (1):\penalty0 1303--1347, 2013.

\bibitem[Ialongo et~al.(2019)Ialongo, Van Der~Wilk, Hensman, and
  Rasmussen]{ialongo2019overcoming}
Alessandro~Davide Ialongo, Mark Van Der~Wilk, James Hensman, and Carl~Edward
  Rasmussen.
\newblock Overcoming mean-field approximations in recurrent gaussian process
  models.
\newblock In \emph{International Conference on Machine Learning}, pages
  2931--2940, 2019.

\bibitem[Kalman(1960)]{kalman1960}
Rudolph~Emil Kalman.
\newblock A new approach to linear filtering and prediction problems.
\newblock \emph{Journal of basic Engineering}, 82\penalty0 (1):\penalty0
  35--45, 1960.

\bibitem[Kamthe and Deisenroth(2017)]{kamthe2017}
Sanket Kamthe and Marc~Peter Deisenroth.
\newblock Data-efficient reinforcement learning with probabilistic model
  predictive control.
\newblock \emph{arXiv preprint arXiv:1706.06491}, 2017.

\bibitem[Khasminskii(2012)]{Khasminskii2012Stochastic-1}
Rafail Khasminskii.
\newblock \emph{Stochastic Stability of Differential Equations}.
\newblock Stochastic Modelling and Applied Probability. {Springer-Verlag},
  Berlin Heidelberg, 2 edition, 2012.

\bibitem[Krishnan et~al.(2017)Krishnan, Shalit, and
  Sontag]{krishnan2017structured}
Rahul~G Krishnan, Uri Shalit, and David Sontag.
\newblock Structured inference networks for nonlinear state space models.
\newblock In \emph{Thirty-First AAAI Conference on Artificial Intelligence},
  2017.

\bibitem[Malik et~al.(2019)Malik, Kuleshov, Song, Nemer, Seymour, and
  Ermon]{malik2019calibrated}
Ali Malik, Volodymyr Kuleshov, Jiaming Song, Danny Nemer, Harlan Seymour, and
  Stefano Ermon.
\newblock Calibrated model-based deep reinforcement learning.
\newblock In \emph{International Conference on Machine Learning}, pages
  4314--4323, 2019.

\bibitem[Markovsky and Rapisarda(2008)]{markovsky2008data}
Ivan Markovsky and Paolo Rapisarda.
\newblock Data-driven simulation and control.
\newblock \emph{International Journal of Control}, 81\penalty0 (12):\penalty0
  1946--1959, 2008.

\bibitem[Matthews(2017)]{matthews2017scalable}
Alexander Graeme de~Garis Matthews.
\newblock \emph{Scalable Gaussian process inference using variational methods}.
\newblock PhD thesis, University of Cambridge, 2017.

\bibitem[Mattos et~al.(2015)Mattos, Dai, Damianou, Forth, Barreto, and
  Lawrence]{mattos2016}
C{\'e}sar Lincoln~C Mattos, Zhenwen Dai, Andreas Damianou, Jeremy Forth,
  Guilherme~A Barreto, and Neil~D Lawrence.
\newblock Recurrent gaussian processes.
\newblock \emph{arXiv preprint arXiv:1511.06644}, 2015.

\bibitem[McHutchon et~al.(2015)]{mchutchon2015nonlinear}
Andrew~James McHutchon et~al.
\newblock \emph{Nonlinear modelling and control using Gaussian processes}.
\newblock PhD thesis, Citeseer, 2015.

\bibitem[Salimbeni and Deisenroth(2017)]{salimbeni2017}
Hugh Salimbeni and Marc Deisenroth.
\newblock Doubly stochastic variational inference for deep gaussian processes.
\newblock In \emph{Advances in Neural Information Processing Systems}, pages
  4588--4599, 2017.

\bibitem[Soong(1973)]{soong1973}
Tsu~T Soong.
\newblock \emph{Random differential equations in science and engineering}.
\newblock Elsevier, 1973.

\bibitem[Titsias(2009)]{titsias2009variational}
Michalis Titsias.
\newblock Variational learning of inducing variables in sparse gaussian
  processes.
\newblock In \emph{Artificial Intelligence and Statistics}, pages 567--574,
  2009.

\bibitem[Wang et~al.(2006)Wang, Hertzmann, and Fleet]{wang2006gaussian}
Jack Wang, Aaron Hertzmann, and David~J Fleet.
\newblock Gaussian process dynamical models.
\newblock In \emph{Advances in neural information processing systems}, pages
  1441--1448, 2006.

\end{thebibliography}
